\documentclass{article}

\usepackage{arxiv}
\usepackage{float}
\usepackage[utf8]{inputenc} 
\usepackage[T1]{fontenc}    
\usepackage[hidelinks]{hyperref}      
\usepackage{url}            
\usepackage{booktabs}       
\usepackage{amsfonts}       
\usepackage{nicefrac}       
\usepackage{microtype}      
\usepackage{graphicx}
\usepackage[numbers]{natbib}
\usepackage{doi}
\usepackage{amsmath}
\usepackage{amssymb}
\usepackage{subcaption}    
\usepackage{algorithm}
\usepackage{algpseudocode}
\usepackage{tabularx}       
\usepackage{placeins}       
\usepackage[numbers]{natbib}



\title{TinyUDE: Solver-Free Universal Differential Equations on Microcontrollers via Lie–Taylor Jet Matching}

\author{
 \textbf{Pranavanath Balamurali} \\
  Department of Electrical and Computer Engineering\\
  The University of Texas at Austin\\
  \texttt{pb27677@utexas.edu} \\
  \and
 \textbf{Hrishi Kamireddy} \\
  Department of Computer Science\\
  The University of Texas at Austin\\
  \texttt{hk27965@utexas.edu} \\
}

\renewcommand{\undertitle}{Research Article}
\renewcommand{\shorttitle}{Solver-Free UDE Training by Noise-Adaptive Jet Matching}

\begin{document}
\maketitle

\vskip -0.1in
\begin{center}
\begin{minipage}{0.92\linewidth}
    \hrule height 0.8pt
    \vskip 0.8em
    \small
    \centerline{\textbf{\large Abstract}}
    \vskip 0.6em
    Training Universal Differential Equations (UDEs) traditionally relies on backpropagating through numerical ODE solvers, creating memory footprints far exceeding the capabilities of edge microcontrollers. We present Lie--Taylor jet matching, a solver-free training framework that fits a hybrid vector field directly to the first and second time-derivatives of observed system states. These derivatives, the truncated Lie--Taylor jet, are estimated online via Savitzky--Golay filtering, yielding fully analytic gradients without automatic differentiation software. We evaluate whether eliminating the solver compromises accuracy against a conventional baseline (fixed-step RK4 integration, multiple shooting, exact discrete adjoints, Adam) sharing identical dynamics, noise models, network architectures, and metrics. While naive derivative matching degrades under sensor noise, our noise-adaptive mechanisms close and reverse this gap: full-rate phase-shifted sampling, a reservoir buffer, cosine-annealed optimization with weight averaging, on-device noise estimation, and polynomial-misfit quality gating. On a damped pendulum and chaotic double pendulum, our method matches or exceeds baseline accuracy at matched data windows and recovers unmodeled damping coefficients. Across noise levels from $0\%$ to $5\%$, it attains a geometric-mean relative field error of $0.65\times$ that of the baseline within $108$\,kB of static memory, compared with megabytes of solver tape. On an ESP32 microcontroller, the on-device run reaches a field error of $0.0020$ and recovers the damping coefficient to $\hat{c}=0.400$ (true $0.400$) within $61.3$\,kB of static memory and $7.24$\,ms per update ($18.1\%$ duty cycle at $25$\,Hz), confirming real-time on-device training is feasible without a numerical solver.
    \vskip 0.8em
    \hrule height 0.4pt
\end{minipage}
\end{center}
\vskip 0.25in

\section{Introduction}
\label{sec:intro}

\subsection{Motivation \& Problem Statement}
Physical systems operating in field conditions (robotic manipulators, unmanned aerial vehicles, electric motors, remote sensing nodes) frequently deviate from initial nominal models. Mechanical components wear over time, aerodynamic properties change, loads vary, and temperature shifts alter internal friction. Universal Differential Equations (UDEs), differential equations whose right-hand side is partly a neural network trained inside the equation, provide a powerful framework to address these variations by combining known physics, expressed as an analytic vector field $f_{\mathrm{known}}$, with a neural network residual $N_\theta$ designed to capture unmodeled dynamics \citep{rackauckas2021ude}. This hybrid formulation preserves physical interpretability and structural extrapolation while granting the model sufficient capacity to capture unexpected disturbances, drag, and dissipation.

Despite their advantages, the standard training pipeline for UDEs creates significant computational hurdles for edge deployments. Typically, the hybrid system is integrated inside an optimization loop, and gradients are computed by backpropagating through the differential equation solver using unrolled automatic differentiation or continuous adjoint states (the adjoint method obtains parameter gradients by integrating a second, backward-in-time differential equation for the loss sensitivities) \citep{chen2018neuralode}. This is often stabilized over long time horizons using multiple shooting \citep{bock1984multiple}, which splits a long trajectory into short windows that are each re-initialized from data so the solver never integrates far from the observations. This standard workflow assumes workstation-class computational resources, requiring dedicated automatic differentiation libraries, megabytes of intermediate state storage (tapes), and full offline trajectory access. 

In this work, we investigate whether Ordinary Differential Equation (ODE) solvers and intermediate backpropagation tapes can be eliminated entirely. By matching estimated time-derivatives in a streaming framework, we examine how solver-free learning compares to traditional adjoint methods in terms of model accuracy. We demonstrate that when derivative estimates are processed with appropriate statistical filtering, direct derivative matching eliminates the collocation accuracy penalty. Specifically, noise-adaptive Lie--Taylor jet matching outperforms validated adjoint baselines while reducing training memory requirements by roughly two orders of magnitude. Because direct numerical differentiation inherently amplifies measurement noise as $\mathcal{O}(\sigma/\Delta t)$, standard least-squares regression on noisy targets produces biased estimates. The mechanisms introduced in Section~\ref{sec:method} address this by creating statistically independent derivative targets, averaging over streaming samples, operating filters at noise-optimal parameters, and rejecting corrupted data points.

\subsection{Edge Constraints: Memory Limitations and Solver Fragility}
Two primary obstacles restrict traditional UDE training on resource-constrained microcontrollers:

\begin{enumerate}
    \item \textbf{Memory Footprint of Backpropagation.} Differentiating through a $K$-step numerical solver requires storing intermediate solver stages and network activations for every step in the forward pass. In our baseline experiments (Section~\ref{sec:setup}), a modest network with 354 parameters trained using a multiple shooting window of $K=40$ requires $5.9$\,MB of memory tape and peaks at $9.1$\,MB of heap usage. Extending the window to $K=80$ to handle higher noise increases memory usage to $11.8$\,MB. Standard microcontrollers, which typically possess around $320$\,kB of static random-access memory (SRAM), cannot host these operations. Consequently, existing embedded ML frameworks focus almost exclusively on model inference \citep{david2021tensorflow, lin2020mcunet}, while on-device training methods are generally restricted to fine-tuning final classification layers rather than updating dynamical systems \citep{lin2024ondevice}.
    \item \textbf{Numerical Integrator Divergence.} Embedded numerical integration inside an optimization loop tightly couples model parameter updates to solver stability. Temporary parameter excursions during training can cause intermediate trajectories to diverge, leading to unstable gradient steps. Furthermore, computing each gradient requires $4K$ vector-field evaluations in both forward and reverse passes. Removing the solver avoids solver instability but introduces a different challenge: training targets must be estimated directly from noisy sensor data. Without adequate filtering, point-to-point collocation fails under noise, and in our experiments, single-step collocation ($K=1$) degraded to a relative field error of $0.82$ under $5\%$ noise. Therefore, solver-free methods require robust statistical filtering to generate reliable derivative targets.
\end{enumerate}

\subsection{Summary of Contributions}
We propose noise-adaptive Lie--Taylor jet matching and present the following key contributions:
\begin{itemize}
    \item \textbf{Solver-Free Training with Exact Analytic Gradients.} We introduce a solver-free training approach that fits hybrid vector fields to first- and second-order time derivatives $(\dot{x}, \ddot{x})$ derived from Savitzky--Golay (SG) filtering \citep{steiner1964smoothing}, a sliding-window least-squares polynomial fit whose derivatives at the window center serve as smoothed derivative estimates. The second derivative is modeled using the Lie derivative identity $\ddot{x} = J_{F_\theta}F_\theta$, where the Lie derivative is the rate of change of a quantity along the flow of the vector field \citep{isidori1995nonlinear}. Both terms yield closed-form analytic gradients, avoiding the need for automatic differentiation software. We validate these gradients against complex-step differentiation (a technique that recovers derivatives of a real function to machine precision by evaluating it at a small imaginary perturbation, free of the cancellation error of finite differences) across a 32-check suite, achieving agreement to machine precision.
    \item \textbf{On-Device Noise-Adaptive Processing.} We implement five lightweight mechanisms designed for microcontrollers to mitigate derivative noise:
    \begin{enumerate}
        \item \emph{Full-rate phase sampling}: By sliding the filtering window every few clock ticks while maintaining tap spacing $m$, we extract approximately $50\times$ more statistically independent derivative targets from raw sensor streams, requiring only a $4.4$\,kB ring buffer.
        \item \emph{Reservoir buffering}: A $2048$--$4096$ sample buffer ($48$--$96$\,kB) maintained by reservoir sampling, which keeps a uniform random subset of an unbounded stream in fixed memory, replaces a small rolling minibatch and prevents the network from memorizing the noise on a handful of targets.
        \item \emph{Averaged Adam optimization}: We use Adam (adaptive moment estimation, a first-order optimizer with per-parameter step sizes) with a cosine-annealed learning rate (a step size that decays along a half cosine so that updates settle) and Polyak weight averaging implemented as an exponential moving average (EMA) of the iterates, which cancels residual optimization noise \citep{kingma2015adam, polyak1992averaging}, all within $4.2$\,kB of state memory.
        \item \emph{Adaptive filter time-scaling}: An on-device noise estimator computes sensor noise directly from fit residuals ($0.04\%$ empirical error) to set optimal filter time scales according to $\Delta t^\ast \propto \sigma^{1/5}/\omega_c$, while gating second-order derivative contributions based on signal-to-noise ratios.
        \item \emph{Polynomial misfit quality gating}: We reject sampling windows where local polynomial assumptions fail, preventing corrupted derivative targets during fast non-linear state transitions.
    \end{enumerate}
    \item \textbf{Experimental Validation and Performance Evaluation.} We evaluate the approach against an exact discrete adjoint baseline sharing identical plants, noise models, neural networks, and evaluation metrics. Jet matching achieves a geometric-mean field error ratio of $0.65\times$ relative to the reference baseline across noise levels from $0\%$ to $5\%$ ($0.82\times$ against a baseline configured with $K=80$). On a chaotic double pendulum, the proposed method yields lower field error ($0.0116$ vs.\ $0.0137$) and accurately recovers unmodeled damping coefficients ($\hat{c}=0.388$--$0.399$ compared to the true value of $0.40$), all within a static memory allocation of $108$\,kB.
\end{itemize}

\section{Related Work}
\label{sec:related}

\subsection{Universal Differential Equations and Neural ODEs}
Neural ODEs place a neural network on the right-hand side of a differential equation and train it by differentiating through the numerical solution, either by unrolling the solver steps or by integrating a continuous adjoint equation backward in time \citep{chen2018neuralode}. The framework has been extended to irregularly sampled time series \citep{rubanova2019latent}, augmented with auxiliary dimensions to relax topological constraints on the learned flow \citep{dupont2019augmented}, and surveyed comprehensively by a recent study \citep{kidger2022on}. UDEs refine the idea for scientific settings by combining a known mechanistic vector field with a neural term that models only the unknown dynamics \citep{rackauckas2021ude}, retaining interpretability and extrapolation where the physics is trusted. A parallel line builds physical structure into the network itself: Hamiltonian neural networks learn a scalar energy whose symplectic gradient gives the dynamics \citep{greydanus2019hamiltonian}, and Lagrangian formulations do the same through the Euler--Lagrange equations \citep{cranmer2020lagrangian, lutter2019deep}. Physics-informed neural networks (PINNs) instead penalize the residual of the governing equations at collocation points using automatic differentiation of a neural representation of the solution \citep{raissi2019physics, karniadakis2021physics}.

All of these approaches share the training obstacle that motivates this work: the gradient is obtained by reverse-mode differentiation through a numerical integration, whose memory grows with the number of solver steps. The literature has attacked the cost from several directions. Checkpointing and adaptive-checkpoint adjoints trade recomputation for memory \citep{gholami2019anode, zhuang2020adaptive}. The distinction between discretize-then-optimize versus optimize-then-discretize clarifies when a continuous adjoint yield exact gradients for the discrete solver used in practice \citep{onken2020discretize}. Meanwhile, multiple shooting splits long trajectories into short windows, allowing solver-based training to tolerate noisy records \citep{bock1984multiple, turan2022multiple}. Collocation-style training of neural ODEs, which fits the field to smoothed derivative estimates rather than integrating, has been studied as a fast initialization or alternative to adjoint training \citep{roesch2021collocation}. Our baseline (Section~\ref{sec:setup}) is a careful instance of the adjoint recipe: fixed-step Runge--Kutta integration \citep{butcher2008numerical, hairer2006geometric}, multiple shooting, an exact discrete adjoint, and its own shooting-length sweep quantifies the memory that noise robustness costs. Jet matching keeps the UDE decomposition but abandons the solve-inside-the-loss construction entirely. Our chaotic test bed, the damped double pendulum, is a standard benchmark for sensitive dependence on initial conditions \citep{shinbrot1992chaos, strogatz2015nonlinear}.

\subsection{On-Device Machine Learning and Memory Constraints}
Deploying machine learning on microcontroller units (MCUs) has matured into a distinct field, TinyML, with interpreter runtimes that execute quantized networks in tens of kilobytes \citep{david2021tensorflow, warden2020tinyml, ren2021tinyol}. Additionally, hardware-aware architecture search fits large classifiers into microcontroller flash \citep{lin2020mcunet}, and standardized benchmarks for inference latency and energy \citep{banbury2021mlperf}. On-device training is far less developed \citep{dhar2021survey}. Memory-efficient transfer learning freezes the weights and updates only biases and small adapter modules to avoid storing activations \citep{cai2020tinytl}. Sparse-update training under a 256\,kB budget selects which layers and channels may change at all \citep{lin2024ondevice}, and online learning on microcontrollers has been demonstrated for incremental classifier adaptation \citep{warden2020tinyml, ren2021tinyol}. These systems target classification and adapt final layers, and none trains a dynamics model whose gradient passes through an integrator. Derivative-free simultaneous perturbation stochastic approximation (SPSA) \citep{spall1992multivariate, spall1998overview} estimates a gradient from two randomly perturbed loss evaluations and therefore needs no backpropagation tape, but its gradient variance grows with the parameter count, limiting it to very small networks on microcontroller-class hardware. Jet matching is complementary and stronger where sensors stream densely: it retains exact analytic gradients of its loss with no zeroth-order variance penalty, while remaining allocation-free and solver-free, and it scales to the hundreds of parameters where perturbation methods degrade.

\subsection{Derivative Estimation, System Identification, and Collocation}
Fitting a vector field directly to estimated state derivatives, $\dot{x} \approx F_\theta(x)$, is the oldest idea in system identification \citep{balakrishnan2002system, nelles2001nonlinear}. Spline and generalized-smoothing methods estimate the trajectory and its derivatives first and then regress the parameters \citep{varah1982spline, ramsay2007parameter}, symbolic regression distills closed-form laws from derivative data \citep{schmidt2009distilling}, and sparse identification of nonlinear dynamics (SINDy) selects a few terms from a candidate library by sparse regression on numerically differentiated data \citep{brunton2016discovering, brunton2022data}, with extensions to learned coordinates \citep{champion2019data} and implicit dynamics \citep{kaheman2020sindypi}. The known weakness of this entire family is what our baseline's $K=1$ ablation exhibits: numerical differentiation amplifies measurement noise, and least squares against corrupted targets is biased. Regularized differentiation methods \citep{chartrand2011numerical, vanbreugel2020numerical} and Savitzky--Golay filtering \citep{steiner1964smoothing, schafer2011what} control the amplification through local polynomial fits, whose first-derivative bias scales as $\mathcal{O}(\Delta t_{\mathrm{eff}}^{4})$ and noise as $\mathcal{O}(\sigma/\Delta t_{\mathrm{eff}})$ at polynomial order four.

Our work builds on this line in three ways:
\begin{enumerate}
    \item We introduce a second-order jet loss term with exact analytic gradients, using acceleration estimates to constrain the Jacobian of the vector field rather than only its value.
    \item We observe that the standard decimated SG pipeline discards $(m-1)/m$ of the stream and correlates adjacent windows, and we recover the discarded independent samples with full-rate phase sampling at negligible device cost.
    \item We introduce closed-form on-device rules that place the filter at its noise-dependent optimum and reject windows where the polynomial model itself fails. The quality gate is in the spirit of robust statistics \citep{huber1964robust}, but acts on the targets' diagnosable failure mode rather than on loss residuals, a distinction our double-pendulum analysis shows to be decisive because a flexible network fits biased targets with small residuals.
\end{enumerate}

Unlike PINNs, which evaluate differential equations across continuous spatio-temporal grids via automatic differentiation, jet matching processes discrete time-series data using localized filtering and closed-form gradient evaluations. Table~\ref{tab:taxonomy} summarizes the key differences.

\begin{table}[H]
\centering
\caption{Taxonomy of differential equation learning frameworks. Memory figures for the adjoint baseline reflect measured usage, while jet matching figures represent static memory allocations.}
\label{tab:taxonomy}
\small
\begin{tabularx}{\linewidth}{l X X l X}
\toprule
Framework & Optimization Paradigm & Noise Handling & MCU Training & Training Memory\\
\midrule
Adjoint UDE \citep{rackauckas2021ude} & Solver in loss, reverse-mode & Multiple shooting & Non-viable & $5.9$--$11.8$\,MB (tape)\\
PINN \citep{raissi2019physics} & Autodiff on implicit repr. & Loss regularization & Offline only & High (grid-dependent)\\
SINDy \citep{brunton2016discovering} & Direct numerical diff. & Smoothing $+$ sparsity & Feasible (linear) & Low\\
Jet Matching (Ours) & Lie--Taylor jet, analytic grads & Full-rate SG $+$ adaptive gates & Microcontroller & $108$\,kB (static)\\
\bottomrule
\end{tabularx}
\end{table}

\subsection{Gradient Computation, Optimization, and Verification}
The two gradient primitives of Section~\ref{sec:method}, the vector--Jacobian product and the Jacobian--vector product, are the elementary operations of reverse- and forward-mode automatic differentiation respectively \citep{griewank2008evaluating, baydin2018automatic}. Composing reverse mode over a forward-mode pass is the classical technique for exact Hessian-vector, and in this case Jacobian-derivative products without materializing matrices \citep{pearlmutter1994fast}. We verify both our gradients and the baseline's adjoint against complex-step differentiation, which evaluates a real function at a small imaginary perturbation and reads the derivative from the imaginary part, avoiding the subtractive cancellation of finite differences and attaining machine precision \citep{squire1998using, martins2003complex}. On the optimization side, our trainer combines stochastic approximation \citep{sinha1971stochastic} in the form of Adam \citep{kingma2015adam} with a cosine learning-rate schedule \citep{loshchilov2017sgdr} and Polyak--Ruppert iterate averaging \citep{polyak1992averaging}, whose modern weight-space form is known to find flatter minima and generalize better \citep{izmailov2019averaging}. The fixed-memory sample buffer is maintained by reservoir sampling \citep{vitter1985random}, which keeps a uniform random subset of an unbounded stream. Each of these components is standard, and the contribution of this paper is showing that, arranged into a streaming, allocation-free pipeline, they eliminate the accuracy penalty that has historically separated derivative-matching from solver-based training.

\section{Methodology}
\label{sec:method}

\subsection{Problem Formulation and Dynamical Systems}
We consider a state observation stream given by $x(t_k) + \eta_k \in \mathbb{R}^{n}$ sampled at frequency $f_s = 1$\,kHz, where $\eta_k$ represents independent and identically distributed (i.i.d.) Gaussian sensor noise. The true physical system follows $\dot{x} = f^\ast(x)$, which we decompose into:
\begin{equation}
    f^\ast(x) = f_{\mathrm{known}}(x) + \Delta(x),
    \label{eq:plant_split}
\end{equation}
where $f_{\mathrm{known}}$ denotes the modeled physical dynamics and $\Delta$ represents unknown dynamics. The target learning model is structured as:
\begin{equation}
    \dot{x} = F_\theta(x) = f_{\mathrm{known}}(x) + N_\theta(x),
    \label{eq:ude}
\end{equation}
where $N_\theta$ is a multilayer perceptron using $\tanh$ activation functions. The neural network architecture uses a $2{\to}16{\to}16{\to}2$ configuration (354 parameters) for the single pendulum and a $4{\to}16{\to}16{\to}4$ configuration (420 parameters) for the double pendulum. Input and output layers are normalized using scale factors determined during an initial calibration phase.

We evaluate performance on two physical systems:
\begin{enumerate}
    \item \textbf{Damped Single Pendulum:} Defined by state vector $x = [\theta, \omega]^{\!\top}$ with governing equations:
    \begin{equation}
        \dot{\theta} = \omega,
        \label{eq:pend_1}
    \end{equation}
    \begin{equation}
        \dot{\omega} = -\frac{g}{L}\sin\theta - c\,\omega,
        \label{eq:pend_2}
    \end{equation}
    where $g = 9.81~\mathrm{m/s^2}$, $L = 1~\mathrm{m}$, and true damping coefficient $c = 0.40$. The system is initialized from $(\theta_0,\omega_0) = (1.2, 0)$ with periodic random re-excitation forces.
    \item \textbf{Chaotic Damped Double Pendulum:} Defined by state vector $x = [\theta_1, \theta_2, \omega_1, \omega_2]^{\!\top}$, with link masses $m_1 = m_2 = 1$\,kg and lengths $l_1 = l_2 = 1$\,m. Defining the angular difference $\delta = \theta_1 - \theta_2$ and common denominator term $\mathrm{D} = 2m_1 + m_2 - m_2\cos(2\delta)$, the equations of motion are:
    \begin{equation}
        \dot{\theta}_1 = \omega_1,
        \label{eq:dp_th1}
    \end{equation}
    \begin{equation}
        \dot{\theta}_2 = \omega_2,
        \label{eq:dp_th2}
    \end{equation}
    \begin{equation}
        \dot{\omega}_1 = \frac{-g(2m_1{+}m_2)\sin\theta_1 - m_2 g \sin(\theta_1{-}2\theta_2) - 2\sin\delta\, m_2\big(\omega_2^2 l_2 + \omega_1^2 l_1 \cos\delta\big)}{l_1\,\mathrm{D}} - c_1\,\omega_1,
        \label{eq:dp1}
    \end{equation}
    \begin{equation}
        \dot{\omega}_2 = \frac{2\sin\delta\,\big(\omega_1^2 l_1 (m_1{+}m_2) + g(m_1{+}m_2)\cos\theta_1 + \omega_2^2 l_2 m_2 \cos\delta\big)}{l_2\,\mathrm{D}} - c_2\,\omega_2,
        \label{eq:dp2}
    \end{equation}
    with damping parameters $c_1 = c_2 = 0.150$. Initial conditions are set to $(2.0, 2.4, 0, 0)$ within the chaotic regime.
\end{enumerate}

To evaluate model identification, we construct three UDE experimental configurations:
\begin{itemize}
    \item \textbf{Case A:} Damping terms are omitted from $f_{\mathrm{known}}$, requiring $N_\theta$ to learn the full damping effect.
    \item \textbf{Case B:} Damping is included in $f_{\mathrm{known}}$, but gravity is mis-specified as $g_{\mathrm{model}} = 0.90\,g$. Here, the correct learned damping residual is zero.
    \item \textbf{Case C:} Both structural errors are present simultaneously (mis-specified gravity and omitted damping).
\end{itemize}

Model accuracy is quantified using the relative vector field error over state domain $\mathcal{X}$:
\begin{equation}
    e_{\mathrm{field}} = \frac{\lVert F_\theta - f^\ast \rVert_{L_2(\mathcal{X})}}{\lVert f^\ast \rVert_{L_2(\mathcal{X})}},
    \label{eq:field_error}
\end{equation}
evaluated over a $41{\times}41$ grid for the single pendulum and over trajectories visited during evaluation for the double pendulum. Recovered damping coefficients $\hat{c}$ are computed by linear regression of the learned network outputs against state velocities.

\subsection{Lie--Taylor Jet Matching Loss and Analytic Gradients}
For any smooth system following \eqref{eq:ude}, the first and second time derivatives satisfy $\dot{x} = F_\theta(x)$ and $\ddot{x} = J_{F_\theta}(x)\,F_\theta(x)$ via the chain rule. These are the first terms of the Lie--Taylor expansion of the local flow, which is the Taylor series in time whose coefficients are successive Lie derivatives of the state along the vector field \citep{isidori1995nonlinear}. Truncating it at second order gives the jet that the method matches. Jet matching minimizes the discrepancy between these expressions and filtered derivative estimates $(\hat{x}, \hat{\dot{x}}, \hat{\ddot{x}})$:

We define the first-derivative residual vector $r_1 \in \mathbb{R}^n$ as:
\begin{equation}
    r_1 = \frac{\hat{\dot{x}} - F_\theta(\hat{x})}{s_1},
    \label{eq:r1_def}
\end{equation}
where $s_1$ is the calibration Root Mean Square (RMS) scale for the first derivative. Similarly, the second-order model prediction $G_\theta(\hat{x}) \in \mathbb{R}^n$ is defined as:
\begin{equation}
    G_\theta(\hat{x}) = J_{F_\theta}(\hat{x})\,F_\theta(\hat{x}),
    \label{eq:g_def}
\end{equation}
and the second-derivative residual vector $r_2 \in \mathbb{R}^n$ is given by:
\begin{equation}
    r_2 = \frac{\hat{\ddot{x}} - G_\theta(\hat{x})}{s_2},
    \label{eq:r2_def}
\end{equation}
where $s_2$ is the calibration RMS scale for the second derivative. Combining these components yields the complete jet matching loss function:
\begin{equation}
    \mathcal{L}(\theta) = \lambda_1 \lVert r_1 \rVert_2^2 + \lambda_2 \lVert r_2 \rVert_2^2,
    \label{eq:jetloss}
\end{equation}
where $\lambda_1$ and $\lambda_2$ represent scalar loss weights.

Because parameters $\theta$ enter solely through $N_\theta$, the exact per-sample loss gradient can be computed analytically without numerical differential equation solvers. We define the backpropagated cotangent vector $\bar{u} \in \mathbb{R}^n$ as:
\begin{equation}
    \bar{u} = -2\left( \lambda_1\, \frac{r_1}{s_1} + \lambda_2\, J_{F_\theta}(\hat{x})^{\!\top} \frac{r_2}{s_2} \right),
    \label{eq:u_bar}
\end{equation}
and the second-order weight vector $w \in \mathbb{R}^n$ as:
\begin{equation}
    w = -2\,\frac{r_2}{s_2}.
    \label{eq:w_def}
\end{equation}
The full gradient of the loss function with respect to $\theta$ is then:
\begin{equation}
    \nabla_\theta \mathcal{L} = \left(\frac{\partial N_\theta(\hat{x})}{\partial \theta}\right)^{\!\top} \bar{u} + \lambda_2\, \nabla_\theta \big\langle w,\; J_{N_\theta}(\hat{x})\, v \big\rangle \Big|_{\,v = \mathrm{sg}[F_\theta(\hat{x})]},
    \label{eq:grad}
\end{equation}
where $\mathrm{sg}[\cdot]$ denotes a stop-gradient operator holding vector $v$ constant during derivative evaluation. This formulation needs only two primitive operations: a vector--Jacobian product (VJP, a reverse-mode pass that propagates an output cotangent back to the parameters) through the network for the first term, and reverse-mode differentiation over a Jacobian--vector product (JVP, a forward-mode pass that propagates a tangent direction $v$ through the network) for the second term \citep{baydin2018automatic, pearlmutter1994fast}. Both are evaluated within fixed layer buffers, so the $n \times p$ Jacobian $\partial N_\theta / \partial \theta$ is never formed and no dynamic memory is allocated during training. Detailed mathematical derivations are provided in Appendix~\ref{app:derivation}.

\subsection{Streaming Derivative Estimation: Full-Rate Phase Sampling}
Derivative targets are derived from an $11$-tap ($M=5$), $4\mathrm{th}$-order polynomial ($P=4$) Savitzky--Golay filter operating with tap stride $m$. The effective sampling step is given by:
\begin{equation}
    \Delta t_{\mathrm{eff}} = \frac{m}{f_s}.
    \label{eq:dt_eff}
\end{equation}
Using relative tap indices $\tau \in \{-M, \dots, M\}$, we construct the Vandermonde matrix $A \in \mathbb{R}^{(2M+1)\times(P+1)}$ (the design matrix of the local polynomial fit, whose columns are the monomials $\tau^{j}$ evaluated at the tap positions) with elements:
\begin{equation}
    A_{ij} = \tau_i^{\,j}, \qquad i \in \{1, \dots, 2M+1\}, \quad j \in \{0, \dots, P\}.
    \label{eq:vander_matrix}
\end{equation}
The row vectors $D_d$ for derivative order $d \in \{0, 1, 2\}$ are given by:
\begin{equation}
    D_d = d!\,\big[(A^{\!\top}A)^{-1}A^{\!\top}\big]_{d\,:},
    \label{eq:d_rows}
\end{equation}
where index $d\,:$ selects the corresponding row of the pseudoinverse. Given a sampled data vector $\mathbf{y} \in \mathbb{R}^{2M+1}$, state and derivative estimates are calculated as:
\begin{equation}
    \hat{x} = D_0^{\!\top}\mathbf{y},
    \label{eq:x_hat}
\end{equation}
\begin{equation}
    \hat{\dot{x}} = \frac{D_1^{\!\top}\mathbf{y}}{\Delta t_{\mathrm{eff}}},
    \label{eq:dx_hat}
\end{equation}
\begin{equation}
    \hat{\ddot{x}} = \frac{D_2^{\!\top}\mathbf{y}}{\Delta t_{\mathrm{eff}}^{2}}.
    \label{eq:ddx_hat}
\end{equation}

\begin{figure}[H]
    \centering
    \includegraphics[width=0.9\linewidth]{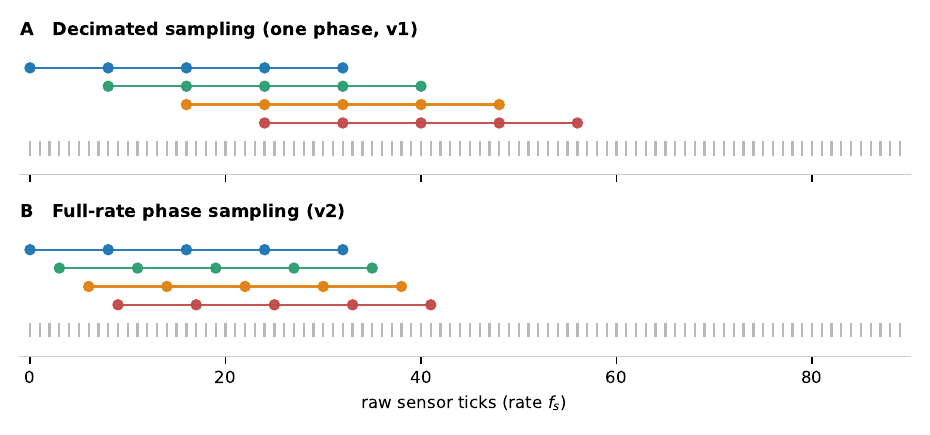}
    \caption{Overview of standard decimated sampling versus full-rate phase sampling. Maintaining tap spacing $m$ while sliding the filter window by stride $s$ yields $m/s$ independent derivative estimation streams.}
    \label{fig:sampling}
\end{figure}

Standard decimation updates filtering windows every $m$ steps, discarding intermediate samples and generating correlated outputs across adjacent windows. To address this, we implement full-rate phase sampling by sliding the filtering window by a small stride $s=5$ while maintaining tap spacing $m$. Because interleaved window phases do not share raw measurement samples, their noise distributions are statistically independent. This increases the effective volume of unique training targets by approximately $m/s$ without requiring additional sensors. The required buffer footprint is limited to $11m$ floating-point values per channel ($4.4$\,kB for single pendulum experiments at $m=50$). Figure~\ref{fig:sampling} illustrates this sampling approach. Filtered outputs are collected into a reservoir buffer of size $B$ ($B=4096$ for single pendulum, $B=2048$ for double pendulum), from which minibatches of size $64$ are drawn for optimization.

\subsection{Noise-Adaptive Filter Tuning and Quality Gating}
We employ three adaptive mechanisms to automatically adjust filter settings based on operational noise conditions:

\textbf{On-Device Noise Estimation.} Let $D_0$ represent the smoothing row vector and $\mu$ denote the center tap index. For a signal corrupted by zero-mean Gaussian noise with variance $\sigma^2$, the scalar residual $e = y_\mu - D_0^{\!\top}\mathbf{y}$ has expected variance $\sigma^2(1 - 2[D_0]_\mu + \lVert D_0 \rVert_2^2)$. Averaging squared residuals over $W$ sliding windows yields an estimator for noise variance:
\begin{equation}
    \hat{\sigma}^2 = \frac{\frac{1}{W}\sum_{k=1}^{W} e_k^{\,2}}{1 - 2\,[D_0]_\mu + \lVert D_0 \rVert_2^2},
    \label{eq:sigmahat}
\end{equation}
which incurs minimal overhead (one multiply-accumulate operation per tap).

\textbf{Adaptive Time-Scale Selection.} Balancing polynomial approximation bias $\mathcal{O}(\Delta t_{\mathrm{eff}}^4)$ against noise variance $\mathcal{O}(\sigma/\Delta t_{\mathrm{eff}})$ yields an optimal time step $\Delta t_{\mathrm{eff}}^{\ast} \propto \sigma^{1/5}$. Scaling this relationship inversely with the characteristic signal frequency $\omega_c = s_2/s_1$ gives:
\begin{equation}
    \Delta t_{\mathrm{eff}}^{\ast} = \frac{0.09}{\omega_c}\left(\frac{\hat{\sigma}_{\mathrm{rel}}}{0.01}\right)^{1/5},
    \label{eq:dt_opt}
\end{equation}
where $\hat{\sigma}_{\mathrm{rel}}$ is the estimated relative noise standard deviation. The constant $0.09$ serves as a baseline scaling factor determined from single pendulum calibration experiments.

\textbf{Second-Order Derivative Gating.} Second derivative estimates $\hat{\ddot{x}}$ carry higher noise standard deviations ($\lVert D_2 \rVert_2 \hat{\sigma}/\Delta t_{\mathrm{eff}}^2$). To prevent noise amplification from degrading optimization, we compute the channel Signal-to-Noise Ratio (SNR):
\begin{equation}
    \mathrm{SNR}_2 = \min_{\text{channels}} \frac{s_2}{\lVert D_2 \rVert_2\, \hat{\sigma} / \Delta t_{\mathrm{eff}}^{2}}.
    \label{eq:snr2_def}
\end{equation}
The second-order loss weight $\lambda_2$ is adaptively assigned based on threshold $\tau_{\mathrm{SNR}} = 12$:
\begin{equation}
    \lambda_2 = \begin{cases} \lambda_2^{\,0} & \text{if } \mathrm{SNR}_2 \ge \tau_{\mathrm{SNR}}, \\[2pt] 0 & \text{if } \mathrm{SNR}_2 < \tau_{\mathrm{SNR}}, \end{cases}
    \label{eq:snrgate}
\end{equation}
where $\lambda_2^{\,0} = 0.1$. If $\mathrm{SNR}_2$ drops below the threshold, second-order derivative matching is disabled to protect gradient updates from high variance.

\textbf{Polynomial Misfit Quality Gating.} High-frequency non-linear dynamics can violate local polynomial assumptions during rapid state transitions. To detect these conditions, we form the residual projection matrix $R = I - A A^{+}$, where $A^{+}$ is the pseudoinverse of $A$. The fit residual norm $\lVert R\,\mathbf{y} \rVert_2$ quantifies polynomial deviation. Samples are rejected if they exceed a dynamic threshold:
\begin{equation}
    \lVert R\,\mathbf{y} \rVert_2 > 6 \cdot \mathrm{median}\big(\lVert R\,\mathbf{y} \rVert_2\big) + 0.02 \cdot \mathrm{RMS}(\mathbf{y}),
    \label{eq:qgate}
\end{equation}
where the median is computed over recent windows and the RMS term establishes a baseline floor. Rejected samples are excluded from the training buffer. Algorithm~\ref{alg:v2} outlines the complete training procedure.

\begin{algorithm}[H]
\caption{Noise-Adaptive Lie--Taylor Jet Matching Routine}
\label{alg:v2}
\begin{algorithmic}[1]
\Require Sensor stream $y_k$ at $f_s = 1$\,kHz; baseline model $f_{\mathrm{known}}$; parameters from Table~\ref{tab:hyper}
\Ensure Optimized weights $\theta_{\mathrm{EMA}}$
\State \textbf{Calibration:} Measure scales $x_{\mathrm{rms}}, s_1, s_2$ using SG estimates \eqref{eq:x_hat}--\eqref{eq:ddx_hat}
\State \textbf{Adaptation:} Estimate noise $\hat{\sigma}$ using \eqref{eq:sigmahat}; set $\omega_c \gets s_2/s_1$, $m \gets f_s \Delta t_{\mathrm{eff}}^{\ast}$ via \eqref{eq:dt_opt}, and weight $\lambda_2$ via \eqref{eq:snrgate}
\For{each incoming sample tick $k$}
    \State Store $y_k$ in circular buffer of length $(2M{+}1)m$ per channel
    \If{$k \bmod s == 0$}
        \State Construct window vector $\mathbf{y}$; evaluate estimates $(\hat{x}, \hat{\dot{x}}, \hat{\ddot{x}})$ via \eqref{eq:x_hat}--\eqref{eq:ddx_hat}
        \If{$\lVert R\,\mathbf{y} \rVert_2$ satisfies condition \eqref{eq:qgate} across all channels}
            \State Add sample $(\hat{x}, \hat{\dot{x}}, \hat{\ddot{x}})$ to reservoir buffer of capacity $B$
        \EndIf
    \EndIf
    \If{$k \bmod (f_s / f_{\mathrm{train}}) == 0$ \textbf{and} buffer is filled}
        \State Draw minibatch of 64 samples; evaluate gradient $\nabla_\theta \mathcal{L}$ using \eqref{eq:u_bar}--\eqref{eq:grad}
        \State Apply Adam update with cosine learning rate schedule; clip gradient norm at $5.0$
        \State Update exponential moving average: $\theta_{\mathrm{EMA}} \gets \beta \theta_{\mathrm{EMA}} + (1{-}\beta)\theta$
    \EndIf
\EndFor
\State Continue training updates on stored reservoir buffer until reaching total iteration limit
\State \Return $\theta_{\mathrm{EMA}}$
\end{algorithmic}
\end{algorithm}

\subsection{Static Memory Allocation}
The training algorithm relies entirely on static memory allocations: network parameters, gradient buffers, Adam optimizer states, and EMA copies require $5 \times 354$ floating-point values for the single pendulum. Including the sample reservoir ($B \times 3n$), circular input buffers ($11m$ values/channel), filter coefficients, scale vectors, and projection matrix $R$, total static memory usage is $108.1$\,kB. This memory footprint remains constant throughout execution, avoiding dynamic allocations during online training. Section~\ref{sec:hardware} reports an on-device measurement of this footprint at a reduced reservoir capacity ($B=2048$, the same capacity used for the double pendulum in Table~\ref{tab:hyper}), adopted to fit the target board's available DRAM alongside the Arduino/ESP-IDF runtime.

\section{Experimental Setup}
\label{sec:setup}

\subsection{Adjoint Baseline Configuration}
We compare the proposed method against a standard solver-based UDE training pipeline: fixed-step 4th-order Runge--Kutta (RK4) integration inside the loss function, multiple shooting over $K$-step windows initialized from trajectory data \citep{bock1984multiple}, exact hand-derived discrete adjoint backpropagation across integration steps (reverse-mode differentiation applied to the discretized solver steps, as opposed to the continuous adjoint), and full-batch Adam optimization \citep{kingma2015adam}. The discrete adjoint implementation was validated against complex-step differentiation, matching numerical gradients to within $1.8{\times}10^{-15}$ to $5.2{\times}10^{-13}$ across test cases and shooting window lengths $K \in \{1,3,5,10\}$. The baseline reference configuration ($K=40$, 128 windows, 1,200 iterations, 50\,Hz observation rate) requires a $5.9$\,MB intermediate tape, $9.1$\,MB peak heap usage, and $26.2$ million multiply-accumulate (MAC) operations per iteration.

\subsection{Experimental Control and Data Matching}
Both jet matching and adjoint baseline implementations share common evaluation components: physical dynamics routines, synthetic noise generation, initial conditions, network architectures, weight initializations, and field error calculation code. Experimental conditions are matched across tests using identical dataset durations ($60$\,s for single pendulum, $20$\,s for double pendulum). Sensitivity checks were performed on baseline parameters: increasing the baseline observation sampling rate to 100\,Hz resulted in higher error ($0.0033$ vs.\ $0.0028$ at the reference $50$\,Hz rate, 1\% noise), while increasing the shooting window to $K=80$ doubled tape size to $11.8$\,MB. Single pendulum experiments were evaluated across 3 random seeds, and double pendulum experiments across 4 seeds.

\subsection{Execution Environment and Resource Projections}
Accuracy evaluations were conducted using 64-bit floating-point simulations in Python/NumPy to isolate algorithmic behavior under controlled numerical precision. Microcontroller hardware projections target an ESP32 microcontroller (Xtensa LX6 dual-core processor running at 240\,MHz, 320\,kB internal SRAM, single-precision floating-point unit (FPU)). Memory figures reflect static compiler allocations. Computation times are projected based on counted operations ($184{,}960$ MACs per update cycle) assuming an execution throughput of 3 clock cycles per MAC and 60 cycles per transcendental function call (\texttt{tanhf}), compared against desktop baseline measurements. Section~\ref{sec:hardware} replaces this projection with a direct on-device measurement from the same training algorithm running on physical ESP32 hardware.

\section{Experiments and Results}
\label{sec:results}

\subsection{Accuracy Across Sensor Noise Levels}
Table~\ref{tab:headline} presents relative field error results across varying sensor noise levels. An unadapted jet matching baseline (v1: decimated sampling, minibatch size 32, fixed learning rates, constant $\lambda_2$) produces a relative field error of $0.0090$ at 1\% noise, a $3.2\times$ higher error than the adjoint baseline ($0.0028$). Incorporating the adaptive mechanisms described in Section~\ref{sec:method} (v2) reduces the relative field error to $0.0022$ under identical 60\,s data windows. Across noise levels from 0\% to 5\%, the proposed method achieves lower mean field error than the reference baseline in five out of six evaluated conditions, yielding a geometric-mean error ratio of $0.65\times$ ($0.82\times$ compared to the baseline using $K=80$). Parameter estimation accuracy remains consistent across noise levels, recovering single pendulum damping parameters within $\hat{c} \in [0.388, 0.399]$ (true value $0.40$), while baseline estimates drift to $0.434$ under 5\% noise.

\begin{table}[H]
\centering
\caption{Relative field error ($\mu \pm \sigma$ over 3 seeds), single pendulum Case A, matched 60\,s data window. Bold text indicates lowest error per noise level. $\dagger$: 1--3 seeds. Error ratios $<1.0$ indicate lower relative error for jet matching.}
\label{tab:headline}
\small
\resizebox{\linewidth}{!}{%
\begin{tabular}{lcccccc}
\toprule
Noise & Adjoint Baseline & Baseline ($K=80$) & Jet Matching (v2) & Ratio (Ref.) & Ratio ($K=80$) & Jet $\hat{c}$ (True: 0.40)\\
\midrule
$0\%$   & $0.00133 \pm 0.00013$ & $0.00133 \pm 0.00013$ & $\mathbf{0.00068 \pm 0.00022}$ & 0.51 & 0.51 & 0.399\\
$0.1\%$ & $\mathbf{0.00123 \pm 0.00012}$ & $\mathbf{0.00123 \pm 0.00012}$ & $0.00223 \pm 0.00060$ & 1.81 & 1.81 & 0.394\\
$0.5\%$ & $0.00158 \pm 0.00019$ & $0.00158 \pm 0.00019$ & $\mathbf{0.00155 \pm 0.00022}$ & 0.98 & 0.98 & 0.398\\
$1\%$   & $0.00283 \pm 0.00134$ & $0.00283 \pm 0.00134$ & $\mathbf{0.00218 \pm 0.00017}$ & 0.77 & 0.77 & 0.397\\
$2\%$   & $0.00695 \pm 0.00328$ & $0.00355^{\dagger}$ & $\mathbf{0.00266 \pm 0.00031}$ & 0.38 & 0.75 & 0.395\\
$5\%$   & $0.02477 \pm 0.00925$ & $0.01180^{\dagger}$ & $\mathbf{0.00642 \pm 0.00151}$ & 0.26 & 0.54 & 0.388\\
\midrule
Geometric Mean & & & & \textbf{0.65} & \textbf{0.82} & \\
\bottomrule
\end{tabular}%
}
\end{table}

\begin{figure}[H]
    \centering
    \includegraphics[width=\linewidth]{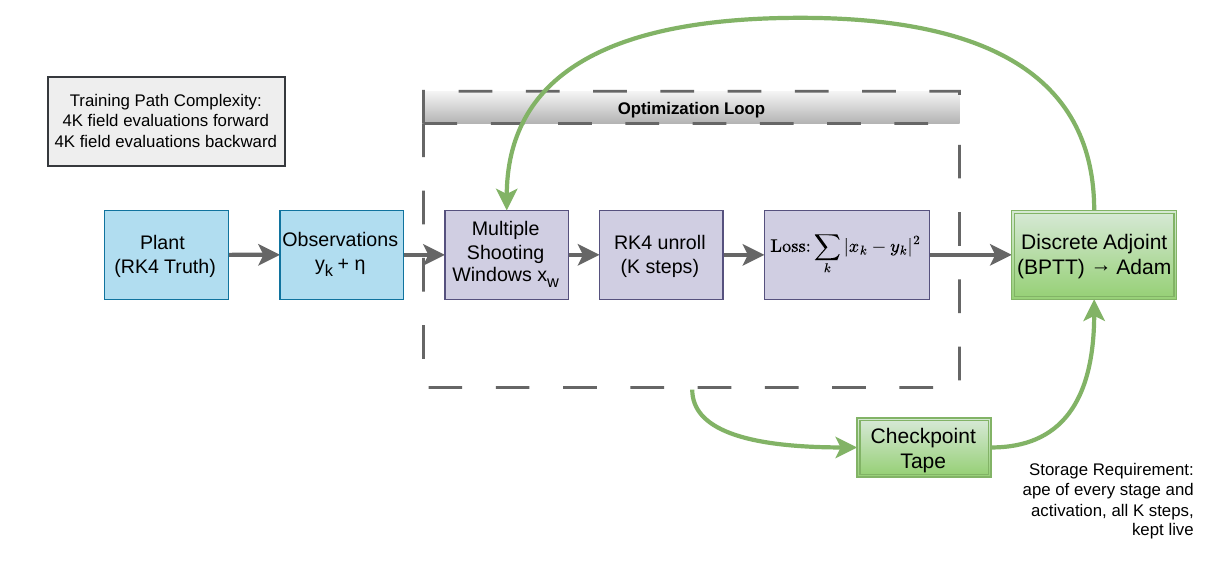}
    \caption{The solver-based training loop that jet matching removes. Each gradient requires $4K$ vector-field evaluations forward and again in reverse, and the checkpoint tape retains every solver stage and activation for all $K$ steps. Jet matching replaces the entire dashed optimization loop and the tape with the streaming pipeline of Algorithm~\ref{alg:v2}.}
    \label{fig:comparison}
\end{figure}

\begin{figure}[H]
    \centering
    \includegraphics[width=\linewidth]{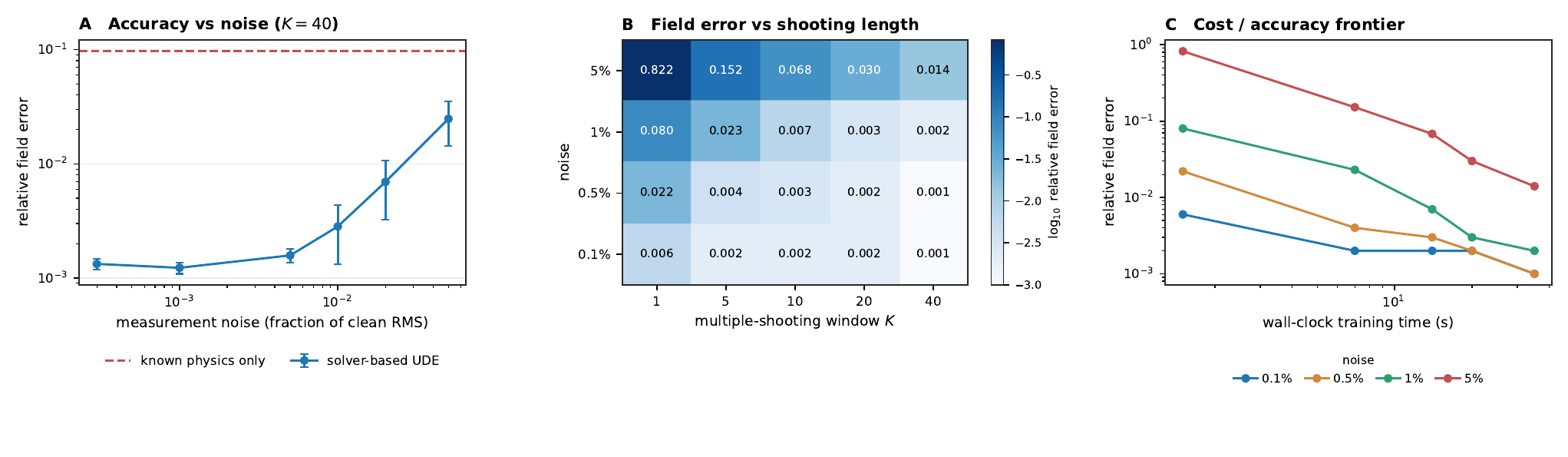}
    \caption{Evaluation of the adjoint baseline: field error versus noise level at $K=40$ (A), accuracy across various shooting window lengths $K$ (B), and trade-offs between computational cost and field accuracy (C).}
    \label{fig:noiseshoot}
\end{figure}

Figure~\ref{fig:comparison} illustrates the solver-based training loop whose solver and tape jet matching removes. Table~\ref{tab:head-to-head} summarizes the operational head-to-head comparison, while Figure~\ref{fig:noiseshoot} illustrates the relationship between shooting window length $K$, memory usage, and noise performance in the baseline model.

\vspace{-15pt}

\begin{table}[H]
\centering
\caption{Detailed comparison under matched physical plants, neural network structures, noise models, and evaluation routines.}
\label{tab:head-to-head}
\small
\begin{tabularx}{\linewidth}{@{} l X X @{}}
\toprule
Metric & Solver-Based Baseline & Jet Matching (Ours) \\
\midrule
Relative field error @ 1\% noise         & 0.0028                     & 0.0022 \\
Relative field error @ 5\% noise         & 0.0248                     & 0.0064 \\
Double pendulum field error @ 1\%        & 0.0137                     & 0.0115 \\
Recovered damping $\hat{c}$ (true: 0.40) & 0.398                      & 0.397 \\
Gradient error vs.\ complex-step         & $1\mathrm{e}{-13}$          & $1\mathrm{e}{-13}$ \\
Numerical ODE solver in loop             & Yes (RK4, 4K evaluations)  & No \\
Backpropagation tape storage             & Yes, $\mathcal{O}(K)$ intermediate states & No \\
Training memory allocation               & 5.9\,MB dynamic tape        & 108\,kB static\textsuperscript{*} \\
Execution time per update step           & 93\,ms (Desktop CPU)        & 2.9\,ms (ESP32 proj.)\textsuperscript{*} \\
Data processing mode                     & Offline trajectory batches  & Online streaming \\
Derivative estimation method             & Solver differentiation      & Adaptive SG filtering \\
\bottomrule
\end{tabularx}
\vskip 2pt
{\small\textsuperscript{*}Projected for the $B=4096$ reference configuration (Table~\ref{tab:hyper}). Section~\ref{sec:hardware} measures the same algorithm on physical ESP32 hardware at a reduced reservoir ($B=2048$, fit to the target board's DRAM): $61.3$\,kB static memory and $7.24$\,ms per update.}
\end{table}

\subsection{Evaluation on Chaotic Double Pendulum Dynamics}
Under 1\% sensor noise over a 20\,s dataset, jet matching achieves a mean relative field error of $0.0115$ (median $0.0083$ across 4 seeds) on the chaotic double pendulum, compared to $0.0137$ for the adjoint baseline (spread $0.0088$ to $0.0198$ across 3 seeds). Evaluating jet matching models on trajectory states generated by the baseline yields a consistent field error of $0.0116$ (median $0.0078$). The model recovers unmodeled damping coefficients at $\hat{c}_1 = 0.141$ (true value $0.150$), compared to the baseline's estimate of $0.160$. 

Occasional error outliers were observed on this system, where 1 out of 4 seeds produced an error of $0.025$. Analysis indicates that during rapid chaotic flips, higher-order Taylor series terms exceed the approximation capacity of 4th-order polynomial filters over fixed window lengths. This causes local derivative estimation errors that pass through the quality gate, temporarily distorting the learned field in high-velocity states ($\lvert \omega \rvert \gg 0$).
\vspace{-10pt}

\subsection{System Identification and Long-Term Rollout Dynamics}
\begin{figure}[H]
    \centering
    \includegraphics[width=\linewidth]{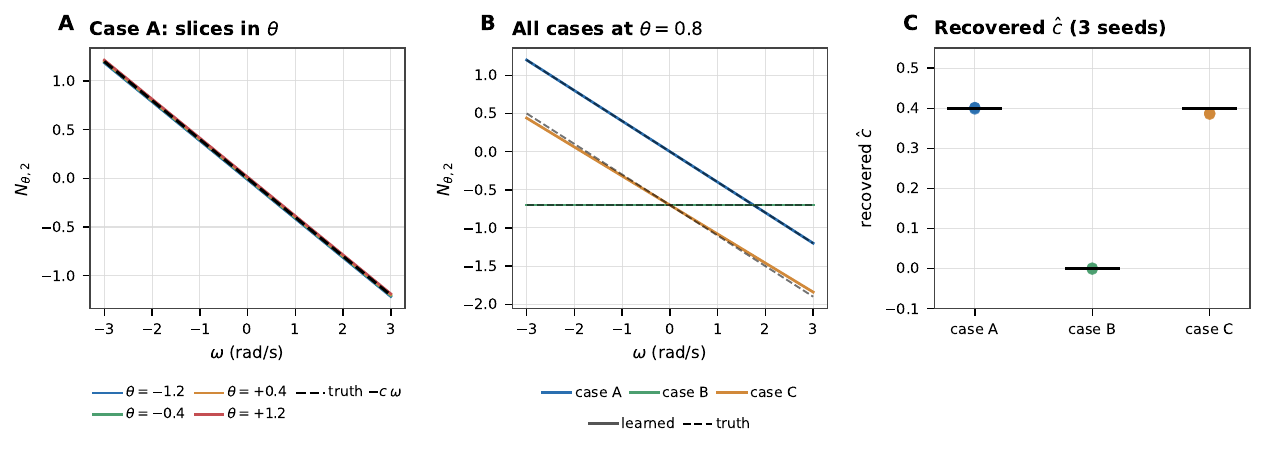}
    \caption{System identification across UDE configurations under 1\% noise: reconstructed residual functions versus true dynamics (A, B) and estimated damping parameters across random seeds (C).}
    \label{fig:ident}
\end{figure}

\begin{figure}[H]
    \centering
    \includegraphics[width=\linewidth]{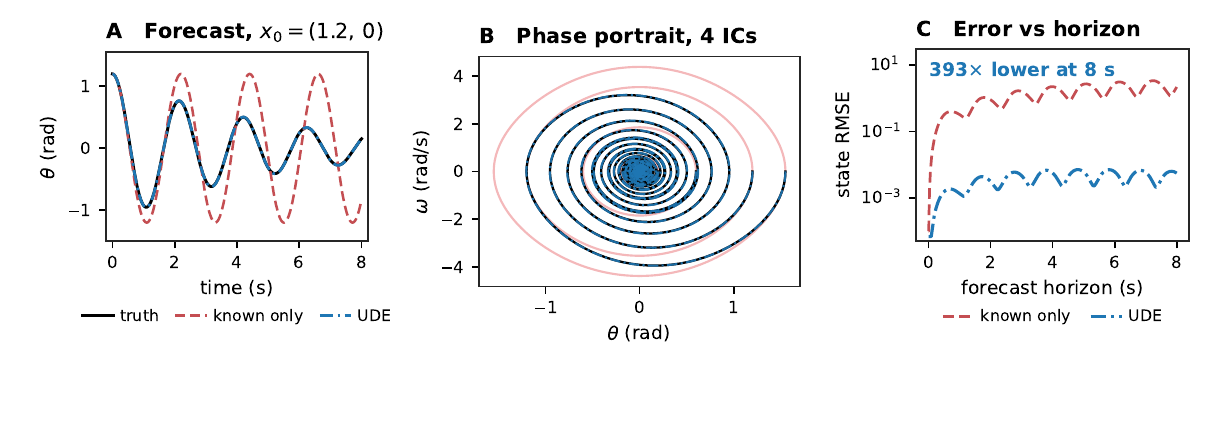}
    \caption{Autonomous trajectory rollouts over an 8\,s horizon (A), phase space trajectories across multiple initial conditions (B), and relative state error progression over time (C).}
    \label{fig:rollout}
\end{figure}

Figure~\ref{fig:ident} illustrates identification performance across different UDE structural formulations under 1\% noise. Case A recovers a damping coefficient of $\hat{c}=0.397$, Case C yields $\hat{c}=0.387$, and Case B, where true damping is fully included in $f_{\mathrm{known}}$, returns an estimated residual coefficient of $-0.001$, confirming that the network does not introduce unnecessary corrections when physical models are accurate. Figure~\ref{fig:rollout} evaluates long-term predictive stability. When integrated over an $8$\,s trajectory window, the trained hybrid field maintains a trajectory error of $0.0068$, representing a $393\times$ reduction in error compared to unaugmented baseline models ($f_{\mathrm{known}}$ alone) and a $10\times$ improvement over unadapted jet matching implementations ($0.068$).

\vspace{-20pt}

\subsection{Ablation Study and Performance Drivers}
Table~\ref{tab:buildup} breaks down performance improvements by adding individual adaptive mechanisms under 1\% noise. The largest accuracy gains stem from full-rate phase sampling and larger sample buffer sizes, which increase sample independence. Dynamic time-scale adaptation and quality gating provide additional refinements, particularly in low-noise settings where fixed filter parameters introduce truncation errors.

\begin{table}[H]
\centering
\caption{Ablation study showing cumulative relative field error improvements on the single pendulum under 1\% noise across 3 random seeds.}
\label{tab:buildup}
\small
\begin{tabularx}{\linewidth}{X l l}
\toprule
Configuration & Field Error & Error vs.\ Baseline ($0.0028$)\\
\midrule
v1: Decimated SG, minibatch size 32, basic SGD & 0.0090 & $3.2\times$ higher\\
$+$ Reservoir buffer ($B=2048$) $+$ Cosine Adam $+$ EMA & 0.0070 & $2.5\times$ higher\\
$+$ Full-rate phase sampling & 0.0037--0.0045 & $1.3$--$1.6\times$ higher\\
$+$ Extended update steps ($8{,}000$) on matched dataset & 0.0023 & $0.81\times$ (lower)\\
$+$ Adaptive filtering, SNR gating, and quality gating (v2) & \textbf{0.0022} & $\mathbf{0.77\times}$ (lower)\\
\bottomrule
\end{tabularx}
\end{table}

\begin{figure}[H]
    \centering
    \includegraphics[width=\linewidth]{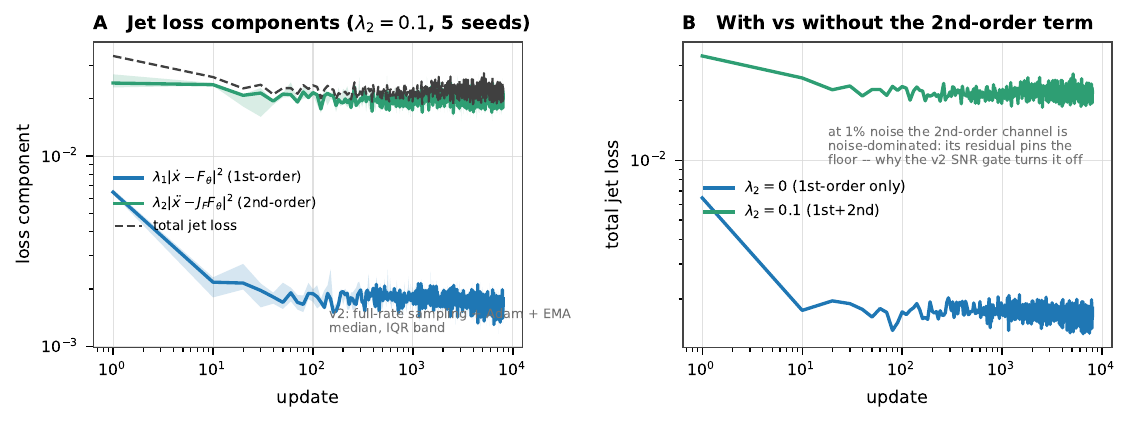}
    \caption{Evaluation of derivative loss components during training at 1\% noise. The second-order loss term remains near a noise floor (A), and disabling it yields equivalent performance (B), supporting the SNR gating strategy.}
    \label{fig:jetcomp}
\end{figure}

\begin{figure}[H]
    \centering
    \includegraphics[width=\linewidth]{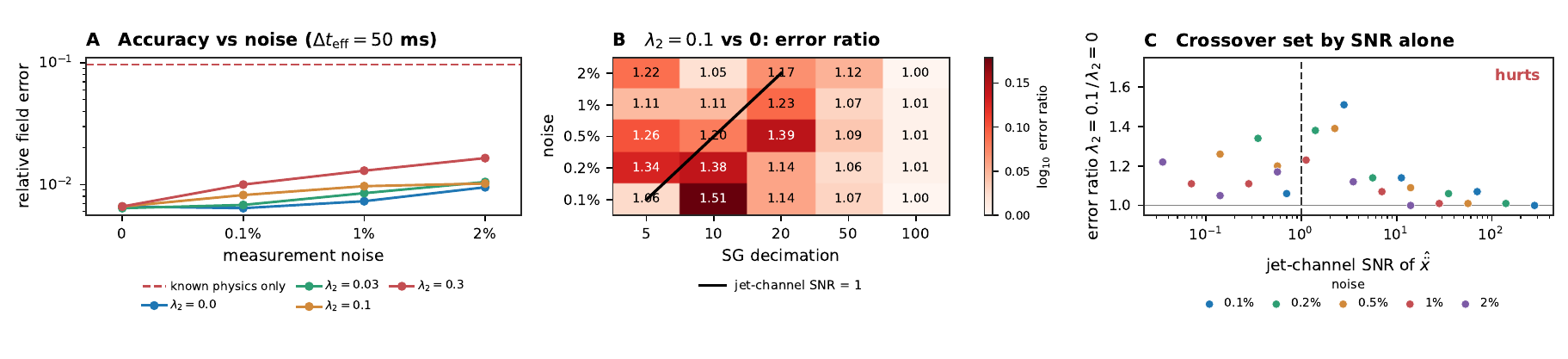}
    \caption{Performance comparison of second-order loss contributions across noise levels and filter decimation steps: field error ratios ($\lambda_2=0.1$ vs.\ $\lambda_2=0$) relative to calculated SNR thresholds.}
    \label{fig:noiselam}
\end{figure}

Figure~\ref{fig:jetcomp} illustrates the impact of second-order derivative gating ($\lambda_2$). Under 1\% noise, the second-order loss component remains pinned near its noise floor ($\sim 2{\times}10^{-2}$) for the entire run, while the first-order loss component drops to $1.8{\times}10^{-3}$. Disabling second-order matching when SNR is low prevents high-variance derivative estimates from impacting parameter updates. Figure~\ref{fig:noiselam} shows performance ratios across different noise levels and filter settings, confirming that second-order matching provides minimal benefit when first-order sample density is sufficient.
\vspace{-8pt}

\subsection{Resource Usage and Memory Requirements}
\begin{figure}[H]
    \centering
    \includegraphics[width=\linewidth]{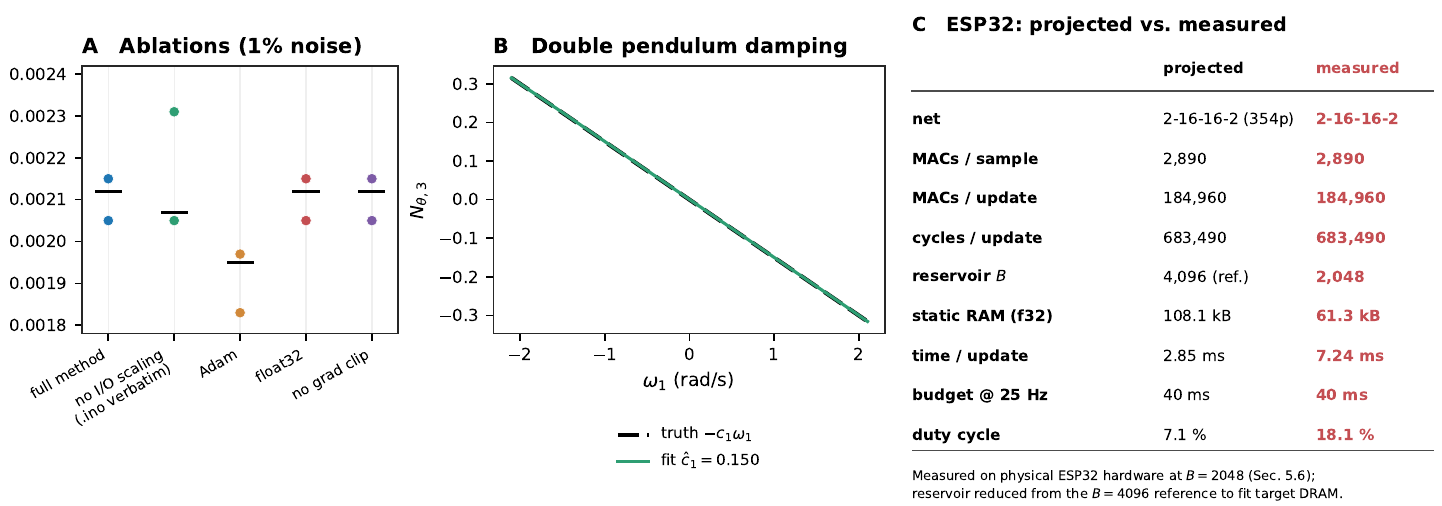}
    \caption{Ablation analysis under 1\% noise (A), noise-free parameter recovery performance (B), and the ESP32 resource account, projected alongside the physical measurement of Section~\ref{sec:hardware} (C).}
    \label{fig:ablcost}
\end{figure}

The baseline model requires $5.9$\,MB of intermediate tape storage ($11.8$\,MB at $K=80$), peaks at $9.1$\,MB of heap allocation, and processes $26.2$\,M MACs per iteration. In contrast, jet matching operates within a static memory allocation of $108.1$\,kB, representing a $55$ to $109\times$ reduction in memory usage. Per-update computational requirements are $184{,}960$ MACs (batch size 64), corresponding to a cycle-counted estimate of $2.85$\,ms per step on a 240\,MHz ESP32 microcontroller ($7.1\%$ duty cycle at 25\,Hz). Figure~\ref{fig:ablcost}C summarizes this projection alongside the corresponding measurement from physical ESP32 hardware, detailed in Section~\ref{sec:hardware}.

\subsection{On-Device Validation on ESP32 Hardware}
\label{sec:hardware}
We implemented the complete training algorithm, including all mechanisms of Section~\ref{sec:method}, in single-precision C on an ESP32 (dual-core Xtensa LX6, 240\,MHz, no floating-point coprocessor for transcendental functions) and ran it end-to-end on physical hardware at $1\%$ sensor noise on the single pendulum. The target board's available DRAM could not accommodate the reference reservoir capacity ($B=4096$, Table~\ref{tab:hyper}) alongside the Arduino/ESP-IDF runtime's own static allocations, so the reservoir was set to $B=2048$, the identical capacity already used for the double pendulum in simulation. The on-device noise estimator measured $\hat{\sigma}_{\mathrm{rel}}=0.0086$ and selected a decimation of $m=26$ (versus $m\approx 39$ in the float64 simulation at the same noise level). Equation~\eqref{eq:dt_opt} depends only on $\omega_c$ and $\hat{\sigma}_{\mathrm{rel}}$, not on reservoir capacity, so this shift reflects estimation variance in $\omega_c$ and $\hat{\sigma}_{\mathrm{rel}}$ over the shorter on-device calibration window together with single-precision arithmetic, rather than the change in $B$ itself. The second-order term remained gated off ($\mathrm{SNR}_2=2.5 < \tau_{\mathrm{SNR}}$), consistent with Section~\ref{sec:setup}.

\begin{figure}[H]
    \centering
    \includegraphics[width=\linewidth]{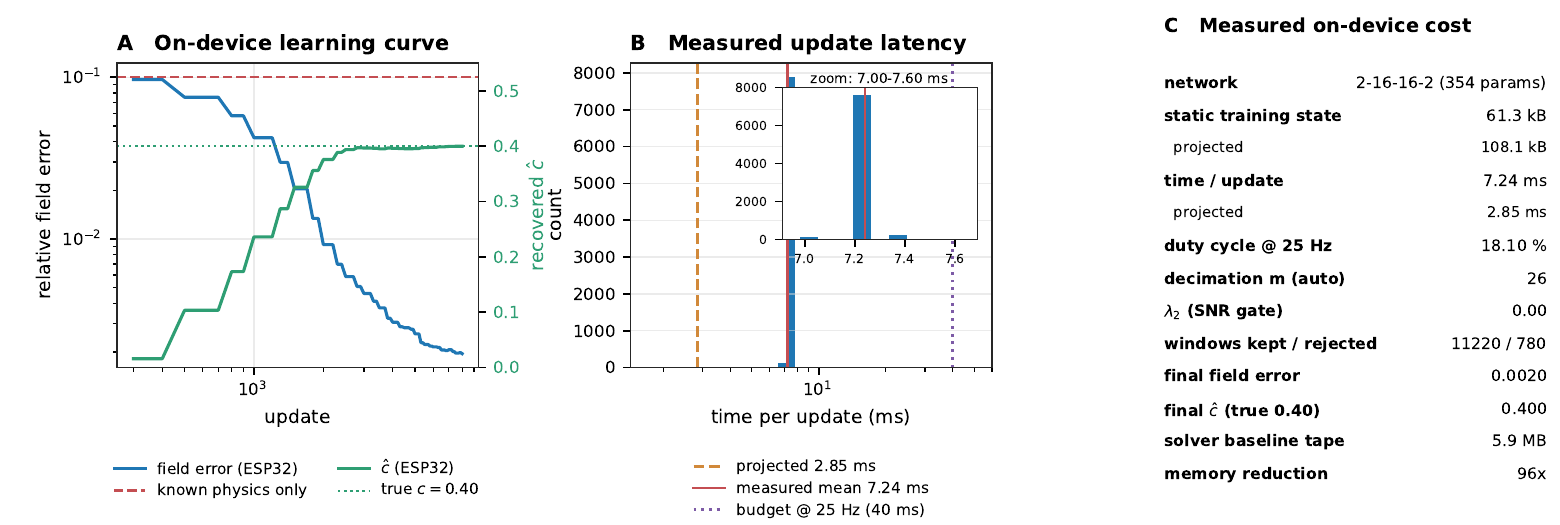}
    \caption{On-device validation on ESP32 hardware at $1\%$ noise, $B=2048$: relative field error and recovered $\hat{c}$ over training (A), measured per-update latency against the projected value and the $25$\,Hz timing budget, with a zoomed inset on the measured spread (B), and the measured resource account against the simulated projection (C).}
    \label{fig:hardware}
\end{figure}

Figure~\ref{fig:hardware} reports the result. The on-device run reaches a relative field error of $0.0020$ and recovers the damping coefficient to $\hat{c}=0.400$ (true $0.400$), matching the accuracy of the float64 simulation at the corresponding operating point despite the smaller reservoir and reduced numerical precision; the quality gate rejected $6.5\%$ of candidate windows ($780$ of $12{,}000$), consistent with the rate reported in Appendix~\ref{app:failure}. Measured static memory usage is $61.3$\,kB, exactly matching the value computed from \texttt{sizeof} on the deployed arrays at $B=2048$, and a $96\times$ reduction relative to the adjoint baseline's $5.9$\,MB tape (consistent with the $55$--$109\times$ range of Section~\ref{sec:results}).

The measured update time is $7.24$\,ms, against the $2.85$\,ms cycle-counted projection of Section~\ref{sec:setup}, a gap we attribute to the projection's idealized cost model: it counts multiply-accumulates and \texttt{tanhf} calls but not the floating-point divisions the algorithm performs per sample (normalizing by $s_x$, $s_1$, $s_2$) or ordinary loop and function-call overhead the compiler does not fully eliminate. The measured time is highly consistent across all $8{,}000$ updates of the run (interquartile range under $0.2$\,ms; inset of Figure~\ref{fig:hardware}B), indicating that the gap is a fixed constant-factor cost rather than variable jitter. At $7.24$\,ms per update the training loop occupies an $18.1\%$ duty cycle at the paper's $25$\,Hz training rate, still comfortably within budget. We report the measured value as the operative timing figure and retain the projection only for its decomposition into MAC and transcendental-call counts.

\section{Conclusion}
\label{sec:conclusion}

\subsection{Summary}
This work evaluates whether numerical differential equation solvers can be removed from UDE training pipelines without sacrificing model accuracy. By replacing internal solver integrations with noise-adaptive Lie--Taylor jet matching, we achieve comparable or superior vector field accuracy across multiple test systems while reducing training memory requirements by roughly two orders of magnitude ($55$ to $109\times$). Full-rate phase sampling, reservoir buffering, adaptive filter tuning, and quality gating allow direct derivative matching to maintain robust performance under measurement noise. The resulting approach operates within $108$\,kB of statically allocated memory, enabling online physics-informed model learning on edge microcontrollers.

\subsection{Limitations and Future Work}
We highlight several operational constraints of the current approach:
\begin{enumerate}
    \item We validated the complete algorithm on physical ESP32 hardware (Section~\ref{sec:hardware}) at a single operating point ($1\%$ noise, single pendulum, $B=2048$). Extending on-device measurement across the full noise sweep, to the double pendulum, and to power consumption remains future work.
    \item In low-noise scenarios ($0.1\%$ noise), traditional adjoint methods maintain an accuracy advantage ($1.8\times$ lower absolute error).
    \item On complex chaotic systems, rapid state transitions can occasionally exceed polynomial filter assumptions, requiring further refinements such as variable window lengths or dynamic filter ordering.
    \item Testing has focused on simulated physical systems under Gaussian noise assumptions, so evaluating performance on hardware platforms with non-Gaussian noise, sensor latency, and quantization effects remains future work.
\end{enumerate}

\section{Acknowledgments}
The authors acknowledge the use of AI tools during the preparation of this manuscript for assistance with rephrasing and paraphrasing text. All AI-generated content was thoroughly reviewed, verified, and edited by the authors, who take full responsibility for the accuracy and integrity of the work presented.

\bibliographystyle{unsrtnat}
\bibliography{references}

\appendix

\section{Derivations}
\label{app:derivation}

\subsection{Reverse-over-JVP Derivation for Second-Order Jet Gradients}
The second-order derivative loss term is given by:
\begin{equation}
    \mathcal{L}_2 = \lambda_2 \lVert r_2 \rVert_2^2,
    \label{eq:l2_def}
\end{equation}
where residual $r_2$ and predicted second derivative $G_\theta$ are defined as:
\begin{equation}
    r_2 = \frac{\hat{\ddot{x}} - G_\theta(\hat{x})}{s_2},
    \label{eq:app_r2}
\end{equation}
\begin{equation}
    G_\theta(\hat{x}) = J_{F_\theta}(\hat{x})\,F_\theta(\hat{x}).
    \label{eq:app_g}
\end{equation}
Differentiating $\mathcal{L}_2$ with respect to parameters $\theta$ yields:
\begin{equation}
    \mathrm{d}\mathcal{L}_2 = -2\lambda_2 \left(\frac{r_2}{s_2}\right)^{\!\top} \mathrm{d}G_\theta,
    \label{eq:dl2_1}
\end{equation}
where the differential $\mathrm{d}G_\theta$ expands to:
\begin{equation}
    \mathrm{d}G_\theta = (\mathrm{d}J_{N_\theta})\,F_\theta(\hat{x}) + J_{F_\theta}(\hat{x})\,\mathrm{d}N_\theta.
    \label{eq:dg_expand}
\end{equation}
Substituting weight vector $w = -2 (r_2 / s_2)$, we separate the gradient into two parts:
\begin{equation}
    \mathrm{d}\mathcal{L}_2 = \lambda_2 \left[ w^{\!\top} (\mathrm{d}J_{N_\theta}) v + w^{\!\top} J_{F_\theta}(\hat{x})\,\mathrm{d}N_\theta \right],
    \label{eq:dl2_split}
\end{equation}
where $v = F_\theta(\hat{x})$. Using the vectorization identity $a^{\!\top} X b = \mathrm{vec}(X)^{\!\top}(b \otimes a)$, the first term transforms to:
\begin{equation}
    w^{\!\top} (\mathrm{d}J_{N_\theta}) v = \mathrm{vec}(\mathrm{d}J_{N_\theta})^{\!\top} (v \otimes w).
    \label{eq:vec_kron}
\end{equation}
This allows computing parameter gradients by reverse-mode differentiation over the Jacobian--vector product $J_{N_\theta} v$ (the forward-mode tangent pass) without explicitly forming higher-order Jacobian matrices.

\subsection{Noise Estimator Variance Identity}
Let a local signal window be represented as $\mathbf{y} = \mathbf{s} + \boldsymbol{\eta}$, where noise $\boldsymbol{\eta} \sim \mathcal{N}(0, \sigma^2 I)$ and $\mathbf{s} = A\mathbf{c}$ represents an exact polynomial sequence. Because Savitzky--Golay smoothing taps $D_0$ preserve polynomials at the center index $\mu$, we have $D_0^{\!\top} A \mathbf{c} = s_\mu$. The scalar fit residual is:
\begin{equation}
    e = y_\mu - D_0^{\!\top}\mathbf{y} = \eta_\mu - D_0^{\!\top}\boldsymbol{\eta}.
    \label{eq:residual_noise}
\end{equation}
Taking the variance yields:
\begin{equation}
    \mathrm{Var}(e) = \mathrm{Var}(\eta_\mu) - 2\mathrm{Cov}(\eta_\mu, D_0^{\!\top}\boldsymbol{\eta}) + \mathrm{Var}(D_0^{\!\top}\boldsymbol{\eta}).
    \label{eq:var_e_expand}
\end{equation}
Substituting component variances $\mathrm{Var}(\eta_\mu) = \sigma^2$, $\mathrm{Cov}(\eta_\mu, D_0^{\!\top}\boldsymbol{\eta}) = \sigma^2 [D_0]_\mu$, and $\mathrm{Var}(D_0^{\!\top}\boldsymbol{\eta}) = \sigma^2 \lVert D_0 \rVert_2^2$ gives:
\begin{equation}
    \mathrm{Var}(e) = \sigma^2 \left(1 - 2[D_0]_\mu + \lVert D_0 \rVert_2^2\right).
    \label{eq:var_e_final}
\end{equation}
Inverting this expression provides the noise variance estimator in \eqref{eq:sigmahat}.

\subsection{Derivation of Damped Single Pendulum Dynamics}
\label{app:damped_pendulum_derivation}

Consider a pendulum of mass $m$ attached to a rigid rod of length $L$ pivoted at the origin $(0,0)$, with angular displacement $\theta$ measured relative to the downward vertical axis. The Cartesian position vector of the pendulum bob is $\vec{r} = (L\sin\theta)\hat{i} + (-L\cos\theta)\hat{j}$. Taking the time derivative gives the velocity vector $\vec{v} = \dot{x}\hat{i} + \dot{y}\hat{j}$, where application of the chain rule yields:
\begin{align}
\dot{x} &= \frac{d}{dt}\left(L \sin\theta\right) = L\dot{\theta}\cos\theta, \\
\dot{y} &= \frac{d}{dt}\left(-L \cos\theta\right) = L\dot{\theta}\sin\theta.
\end{align}

The magnitude squared of the velocity vector is $\|\vec{v}\|^2 = \dot{x}^2 + \dot{y}^2$. Substituting these velocity components into the kinetic energy formulation $T$ and applying the identity $\cos^2\theta + \sin^2\theta = 1$ gives:
\begin{equation}
T = \frac{1}{2}m(\dot{x}^2 + \dot{y}^2) = \frac{1}{2}m\left(L^2\dot{\theta}^2\cos^2\theta + L^2\dot{\theta}^2\sin^2\theta\right) = \frac{1}{2}mL^2\dot{\theta}^2.
\end{equation}
Taking $y=0$ as the reference plane, the potential energy is $V = mgy = -mgL\cos\theta$. The system Lagrangian $L_g = T - V$ is thus formulated as:
\begin{equation}
L_g = \frac{1}{2}mL^2\dot{\theta}^2 + mgL\cos\theta.
\end{equation}

To model mechanical dissipation, we introduce a generalized non-conservative viscous damping force $Q_{\text{nc}} = -b\dot{\theta}$, where $b$ is the rotational damping coefficient. Applying the generalized Euler-Lagrange equation with dissipation:
\begin{equation}
\frac{d}{dt}\left(\frac{\partial L_g}{\partial \dot{\theta}}\right) - \frac{\partial L_g}{\partial \theta} = Q_{\text{nc}}.
\end{equation}

Evaluating the partial derivatives yields $\frac{\partial L_g}{\partial \dot{\theta}} = mL^2\dot{\theta} \implies \frac{d}{dt}\left(\frac{\partial L_g}{\partial \dot{\theta}}\right) = mL^2\ddot{\theta}$ and $\frac{\partial L_g}{\partial \theta} = -mgL\sin\theta$. Substituting these terms into the Euler-Lagrange equation produces:
\begin{equation}
mL^2\ddot{\theta} - (-mgL\sin\theta) = -b\dot{\theta} \implies \ddot{\theta} + \frac{g}{L}\sin\theta = -\frac{b}{mL^2}\dot{\theta}.
\end{equation}

Defining state variables $x_1 = \theta$ and $x_2 = \omega = \dot{\theta}$, alongside the normalized damping factor $c = \frac{b}{mL^2}$, reduces the second-order differential equation to the first-order autonomous state-space system:
\begin{align}
\dot{\theta} &= \omega, \\
\dot{\omega} &= -\frac{g}{L}\sin\theta - c\omega.
\end{align}

\subsection{Derivation of Damped Double Pendulum Dynamics}
\label{app:double_pendulum_derivation}

Consider a double pendulum composed of two point masses $m_1$ and $m_2$ suspended by rigid, massless rods of lengths $l_1$ and $l_2$, respectively. Let $\theta_1$ and $\theta_2$ denote the angular displacements of the upper and lower links relative to the downward vertical axis.

The Cartesian coordinates of $m_1$ and $m_2$ are given by:
\begin{align}
x_1 &= l_1\sin\theta_1, & y_1 &= -l_1\cos\theta_1 \\
x_2 &= l_1\sin\theta_1 + l_2\sin\theta_2, & y_2 &= -l_1\cos\theta_1 - l_2\cos\theta_2
\end{align}

Differentiating with respect to time $t$ yields the linear velocity components $(\dot{x}_1, \dot{y}_1)$ and $(\dot{x}_2, \dot{y}_2)$. Substituting these components into the total kinetic energy expression $T = \frac{1}{2}m_1(\dot{x}_1^2 + \dot{y}_1^2) + \frac{1}{2}m_2(\dot{x}_2^2 + \dot{y}_2^2)$ and applying trigonometric angle addition identities produces:
\begin{equation}
T = \frac{1}{2}(m_1 + m_2)l_1^2\dot{\theta}_1^2 + \frac{1}{2}m_2 l_2^2\dot{\theta}_2^2 + m_2 l_1 l_2 \dot{\theta}_1 \dot{\theta}_2 \cos(\theta_1 - \theta_2)
\end{equation}

Taking the fixed pivot at the origin $(0,0)$ as the potential energy reference plane ($y=0$), the potential energy $V$ is:
\begin{equation}
V = m_1 g y_1 + m_2 g y_2 = -(m_1 + m_2)g l_1 \cos\theta_1 - m_2 g l_2 \cos\theta_2
\end{equation}

Formulating the system Lagrangian $L_g = T - V$ and introducing Rayleigh dissipation forces $Q_{\text{nc}, 1} = -c_1 \dot{\theta}_1$ and $Q_{\text{nc}, 2} = -c_2 \dot{\theta}_2$, we evaluate the coupled Euler-Lagrange equations $\frac{d}{dt}\left(\frac{\partial L_g}{\partial \dot{\theta}_i}\right) - \frac{\partial L_g}{\partial \theta_i} = Q_{\text{nc}, i}$ for $i \in \{1, 2\}$. 

Defining angular velocities $\omega_1 = \dot{\theta}_1$, $\omega_2 = \dot{\theta}_2$, and the angular difference $\delta = \theta_1 - \theta_2$, the resulting system can be expressed as a linear matrix equation in angular accelerations $(\dot{\omega}_1, \dot{\omega}_2)$:
\begin{equation}
\begin{bmatrix}
(m_1 + m_2)l_1 & m_2 l_2 \cos\delta \\
m_2 l_1 \cos\delta & m_2 l_2
\end{bmatrix}
\begin{bmatrix}
\dot{\omega}_1 \\
\dot{\omega}_2
\end{bmatrix}
=
\begin{bmatrix}
-m_2 l_2 \omega_2^2 \sin\delta - (m_1 + m_2)g\sin\theta_1 - c_1 \omega_1 \\
m_2 l_1 \omega_1^2 \sin\delta - m_2 g \sin\theta_2 - c_2 \omega_2
\end{bmatrix}
\end{equation}

Solving the $2 \times 2$ linear system via Cramer's Rule requires the determinant of the mass matrix $M$:
\begin{equation}
\det(M) = m_2 l_1 l_2 \left(m_1 + m_2 - m_2 \cos^2\delta\right)
\end{equation}

Using the double-angle identity $\cos^2\delta = \frac{1 + \cos(2\delta)}{2}$, we define the common denominator parameter $D$:
\begin{equation}
D = 2(m_1 + m_2 - m_2 \cos^2\delta) = 2m_1 + m_2 - m_2\cos(2\delta)
\end{equation}

Inverting the matrix system and isolating $\dot{\omega}_1$ and $\dot{\omega}_2$ yields the explicit state-space system equations:
\begin{align}
\dot{\theta}_1 &= \omega_1 \\
\dot{\theta}_2 &= \omega_2 \\
\dot{\omega}_1 &= \frac{-g(2m_1 + m_2)\sin\theta_1 - m_2 g \sin(\theta_1 - 2\theta_2) - 2\sin\delta \, m_2 \left(\omega_2^2 l_2 + \omega_1^2 l_1 \cos\delta\right)}{l_1 D} - c_1 \omega_1 \\
\dot{\omega}_2 &= \frac{2\sin\delta \left(\omega_1^2 l_1 (m_1 + m_2) + g(m_1 + m_2)\cos\theta_1 + \omega_2^2 l_2 m_2 \cos\delta\right)}{l_2 D} - c_2 \omega_2
\end{align}

\section{Hyperparameter Reference}
\label{app:hyper}

Table~\ref{tab:hyper} lists all configuration parameters used across single and double pendulum experiments.

\begin{table}[H]
\centering
\caption{System parameters and configuration settings.}
\label{tab:hyper}
\small
\begin{tabular}{lcc}
\toprule
Parameter & Single Pendulum & Double Pendulum\\
\midrule
State dimension $n$ / Network & $n=2$, $2{\to}16{\to}16{\to}2$ ($p=354$) & $n=4$, $4{\to}16{\to}16{\to}4$ ($p=420$)\\
Sensor rate / Training rate & \multicolumn{2}{c}{$f_s = 1$\,kHz, $f_{\mathrm{train}} = 25$\,Hz, $8{,}000$ updates}\\
SG window / Order / Stride & \multicolumn{2}{c}{$2M{+}1 = 11$, $P = 4$, $s = 5$}\\
Decimation step $m$ & Automatic ($m\approx 39$ at 1\% noise) & Automatic ($m\approx 17$--$18$ at 1\% noise)\\
Reservoir capacity $B$ / Batch & $4096$ / $64$ & $2048$ / $64$\\
Optimizer / Schedule & \multicolumn{2}{c}{Adam ($\text{lr}=0.02$), Cosine decay to 2\%, Norm clip $5.0$}\\
EMA decay rate $\beta$ & \multicolumn{2}{c}{$0.999$}\\
Loss weights $\lambda_1 / \lambda_2^{\,0} / \tau_{\mathrm{SNR}}$ & \multicolumn{2}{c}{$1.0$ / $0.1$ / $12$}\\
Quality gate parameters & \multicolumn{2}{c}{$6 \times \mathrm{median} + 0.02 \times \mathrm{RMS}$, per channel}\\
Dataset duration & $60$\,s & $20$\,s\\
Static memory footprint & $108.1$\,kB & ${\approx}104$\,kB ($B=2048$)\\
\bottomrule
\end{tabular}
\end{table}

\section{Analysis of High-Velocity Error Outliers}
\label{app:failure}

Experimental runs on the double pendulum identified localized error increases during rapid state transitions ($m=24$). In failing seeds, training loss remained low ($5.9{\times}10^{-3}$) while field errors exceeded $0.11$. Evaluating error distributions across state space showed that localized discrepancies occurred primarily at high angular velocities (e.g., $x = [-0.73, -0.30, 4.67, -7.80]$), where high derivative magnitudes caused polynomial filter approximation errors. 

Because neural networks can fit biased targets tightly, standard robust loss functions (such as Huber losses) do not automatically reject these points. Quality gating via polynomial misfit verification \eqref{eq:qgate} detects these violations directly from raw sensor windows, reducing peak field errors from $0.135$ to $0.025$.

\section{Extended Per-Seed Results}
\label{app:data}

Table~\ref{tab:perseed} provides detailed numerical results across individual random seeds.

\begin{table}[H]
\centering
\caption{Individual per-seed relative field error values across test configurations.}
\label{tab:perseed}
\small
\begin{tabularx}{\linewidth}{X l l}
\toprule
Test Case & Individual Seed Values & Mean\\
\midrule
Jet v2, $0\%$ noise & 0.00048, 0.00091, 0.00064 & 0.00068\\
Jet v2, $0.1\%$ noise & 0.00163, 0.00284, 0.00222 & 0.00223\\
Jet v2, $0.5\%$ noise & 0.00163, 0.00172, 0.00131 & 0.00155\\
Jet v2, $1\%$ noise & 0.00205, 0.00212, 0.00236 & 0.00218\\
Jet v2, $2\%$ noise & 0.00249, 0.00302, 0.00247 & 0.00266\\
Jet v2, $5\%$ noise & 0.00810, 0.00603, 0.00515 & 0.00642\\
Jet v2, Double Pendulum $1\%$ (Native grid) & 0.0254, 0.0054, 0.0112, 0.0038 & 0.0115\\
Jet v2, Double Pendulum $1\%$ (baseline evaluation sets, mean over 3) & 0.0270, 0.0051, 0.0105, 0.0040 & 0.0116\\
Baseline, Double Pendulum $1\%$ noise & 0.0088, 0.0198, 0.0124 & 0.0137\\
Baseline $K=80$, $5\%$ noise & 0.0096, 0.0059, 0.0198 & 0.0118\\
Baseline $K=80$, $2\%$ noise & 0.0048, 0.0023 & 0.0036\\
Baseline $K=80$, $1\%$ noise & 0.0030 & 0.0030\\
\bottomrule
\end{tabularx}
\end{table}

\section{Supplementary Gradient and Estimator Validation Plots}
\label{app:plots}

This section contains supplementary validation plots for numerical gradient checks, Savitzky--Golay filter characteristics, and baseline sensitivity evaluations.

\begin{figure}[H]
    \centering
    \includegraphics[width=0.85\linewidth]{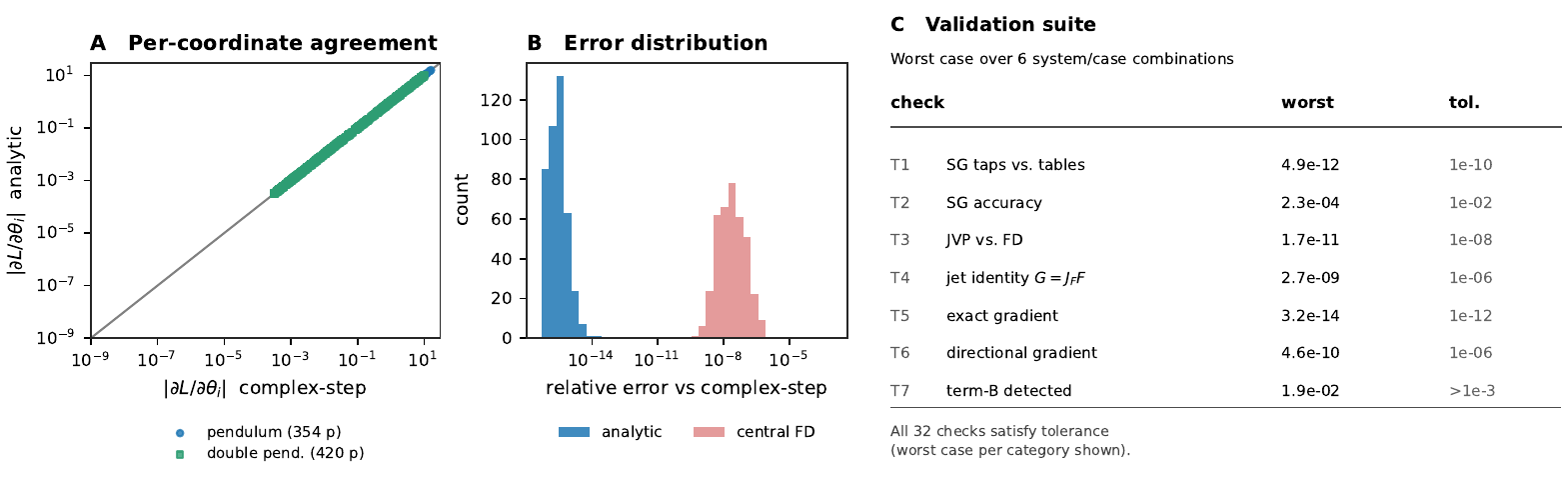}
    \caption{Validation of analytic jet matching gradients against complex-step differentiation across 32 test conditions.}
    \label{fig:gradval_jet}
\end{figure}

\begin{figure}[H]
    \centering
    \includegraphics[width=0.85\linewidth]{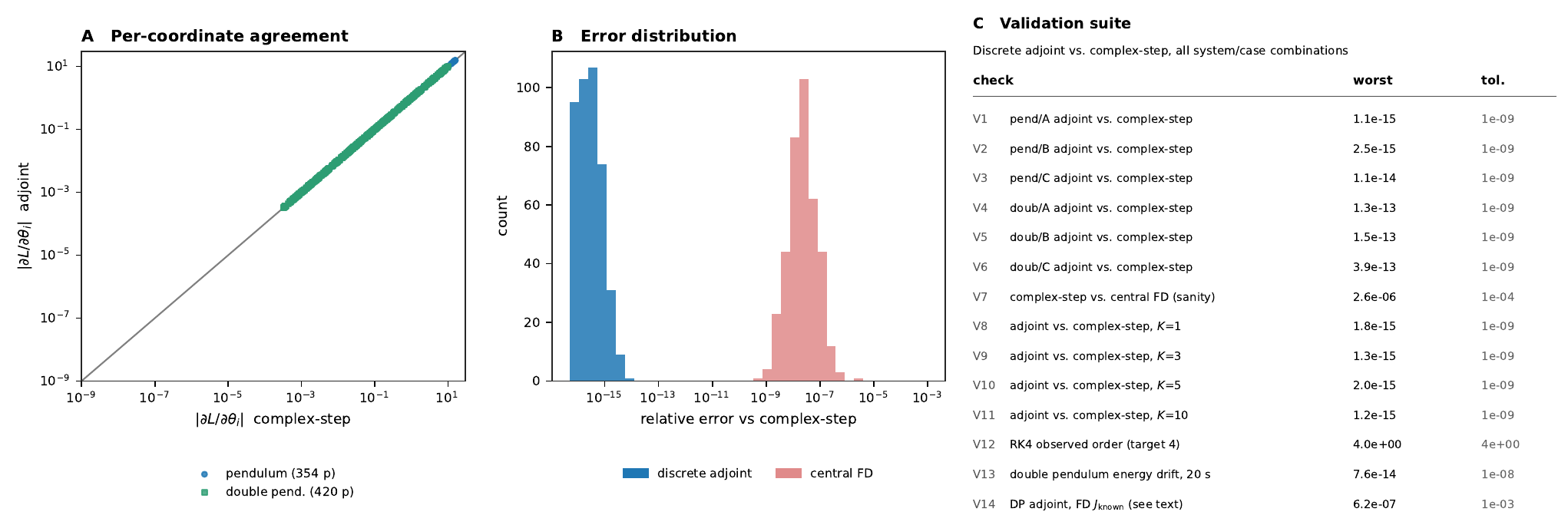}
    \caption{Validation of baseline discrete adjoint gradients against complex-step differentiation across 14 test conditions.}
    \label{fig:gradval_solver}
\end{figure}

\begin{figure}[H]
    \centering
    \includegraphics[width=0.85\linewidth]{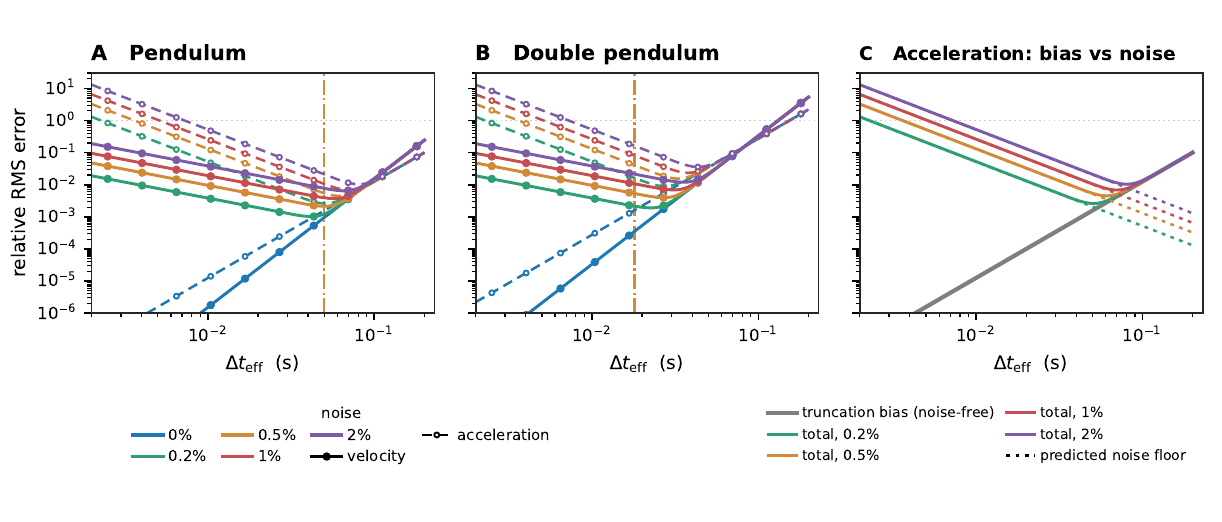}
    \caption{Savitzky--Golay derivative estimation error across effective step sizes $\Delta t_{\mathrm{eff}}$ and noise levels.}
    \label{fig:estimator}
\end{figure}

\begin{figure}[H]
    \centering
    \includegraphics[width=0.85\linewidth]{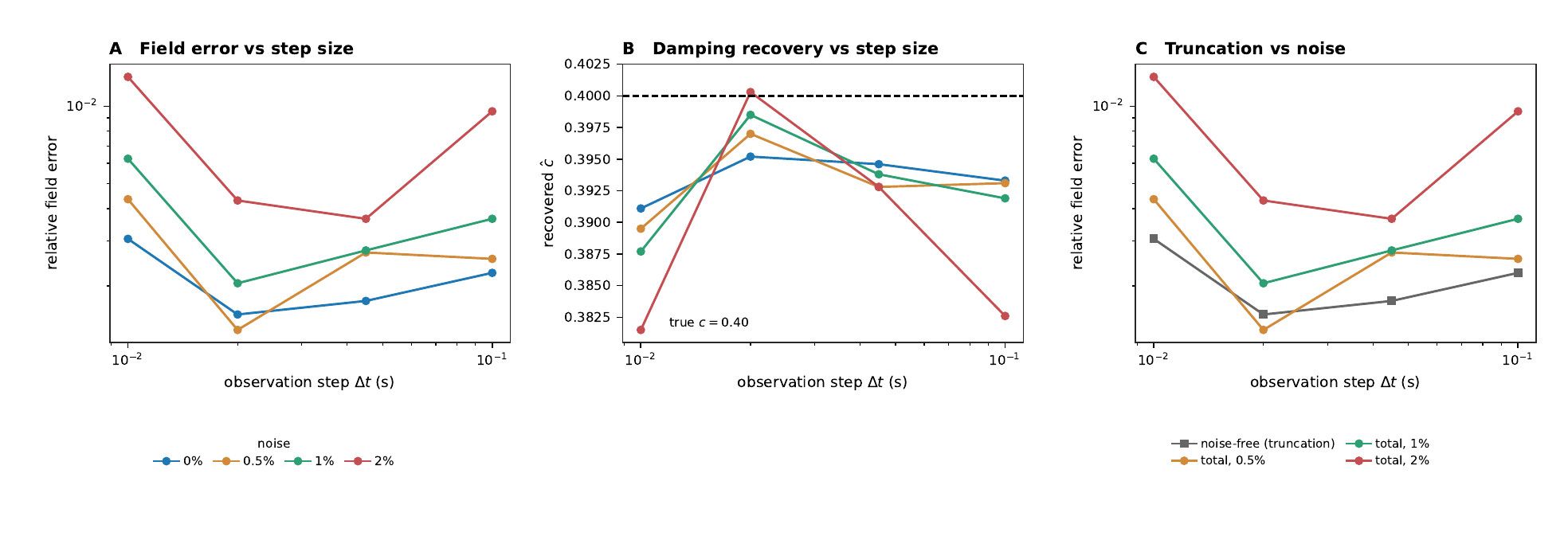}
    \caption{Effect of observation sampling rate on baseline adjoint field error.}
    \label{fig:stepsize}
\end{figure}

\end{document}